\documentclass[letterpaper, 10 pt, conference]{ieeeconf}  
\usepackage[style=numeric]{biblatex}

\IEEEoverridecommandlockouts                              

\title{\LARGE \bf
SLAM in the Dark: Self-Supervised Learning of Pose, Depth and Loop-Closure from Thermal Images
}

\author{Yangfan Xu, Qu Hao, Lilian Zhang, Jun Mao, Xiaofeng He, Wenqi Wu, Changhao Chen*
\thanks{The authors are with the College of Intelligence Science and Technology, National University of Defense Technology, Changsha, 410073, China}
\thanks{Yangfan Xu and Hao Qu contributed equally to this work.}
\thanks{*Corresponding author: Changhao Chen (changhao.chen@nudt.edu.cn).}
}

\usepackage{tikz}
\usepackage[colorlinks=true, linkcolor=blue, urlcolor=blue]{hyperref}
\usepackage{amsmath}
\usepackage{hyperref}
\usepackage{multirow}
\usepackage{booktabs}

\usepackage{biblatex}  
\begin{document}

\maketitle
\thispagestyle{empty}
\pagestyle{empty}
\maketitle

\global\csname @topnum\endcsname 0
\global\csname @botnum\endcsname 0
\thispagestyle{empty}
\pagestyle{empty}

\begin{abstract}
Visual SLAM is essential for mobile robots, drone navigation, and VR/AR, but traditional RGB camera systems struggle in low-light conditions, driving interest in thermal SLAM, which excels in such environments. However, thermal imaging faces challenges like low contrast, high noise, and limited large-scale annotated datasets, restricting the use of deep learning in outdoor scenarios. We present DarkSLAM, a noval deep learning-based monocular thermal SLAM system designed for large-scale localization and reconstruction in complex lighting conditions.Our approach incorporates the Efficient Channel Attention (ECA) mechanism in visual odometry and the Selective Kernel Attention (SKA) mechanism in depth estimation to enhance pose accuracy and mitigate thermal depth degradation. Additionally, the system includes thermal depth-based loop closure detection and pose optimization, ensuring robust performance in low-texture thermal scenes. Extensive outdoor experiments demonstrate that DarkSLAM significantly outperforms existing methods like SC-Sfm-Learner and Shin et al., delivering precise localization and 3D dense mapping even in challenging nighttime environments. 
\end{abstract}

\section{INTRODUCTION}
Simultaneous Localization and Mapping (SLAM) is crucial for intelligent systems from mobile robots, drones, to self-driving vehicles, enabling their real-time localization and mapping for autonomous navigation. Traditional visual SLAM systems, which rely on visible-light (RGB) cameras, struggle in challenging lighting conditions such as strong light, shadows, or nighttime, limiting their use in all-time scenarios. Thermal cameras, which detect heat radiation, offer a solution by functioning in darkness, smoke, and dust, complementing visible-light sensors. Recent improvements in thermal camera resolution and sensitivity have increased their reliability in autonomous systems.

While thermal technology is well-established in areas such as object detection and identification, its application in SLAM remains limited due to the challenges of feature extraction, matching and reconstruction in thermal images. However, thermal SLAM offers significant advantages in low-light environments, where visible-light SLAM often fails. As a result, research on thermal SLAM is expanding, with efforts focused on adapting traditional SLAM frameworks to process thermal data. For example,
Buemi et al. \cite{Buemi_2021_ICCV}, and Li et al.  \cite{li2017line} have developed methods to extract usable features from thermal imagery. Additionally, thermal sensors are increasingly integrated with LiDAR, RGB cameras, and inertial sensors to improve perception and localization in complex environments. Works such as those by Chen    \cite{chen2022eil} and Choi   \cite{CHOI2023222} combine thermal data with LiDAR, while others \cite{9804793} fuse thermal imagery with RGB and inertial data. Despite these advancements, monocular thermal SLAM remains challenging due to the limited visual information and high noise levels in thermal images, and a fully effective solution has yet to be developed.

\begin{figure}
    \centering
    \includegraphics[width=1\columnwidth]{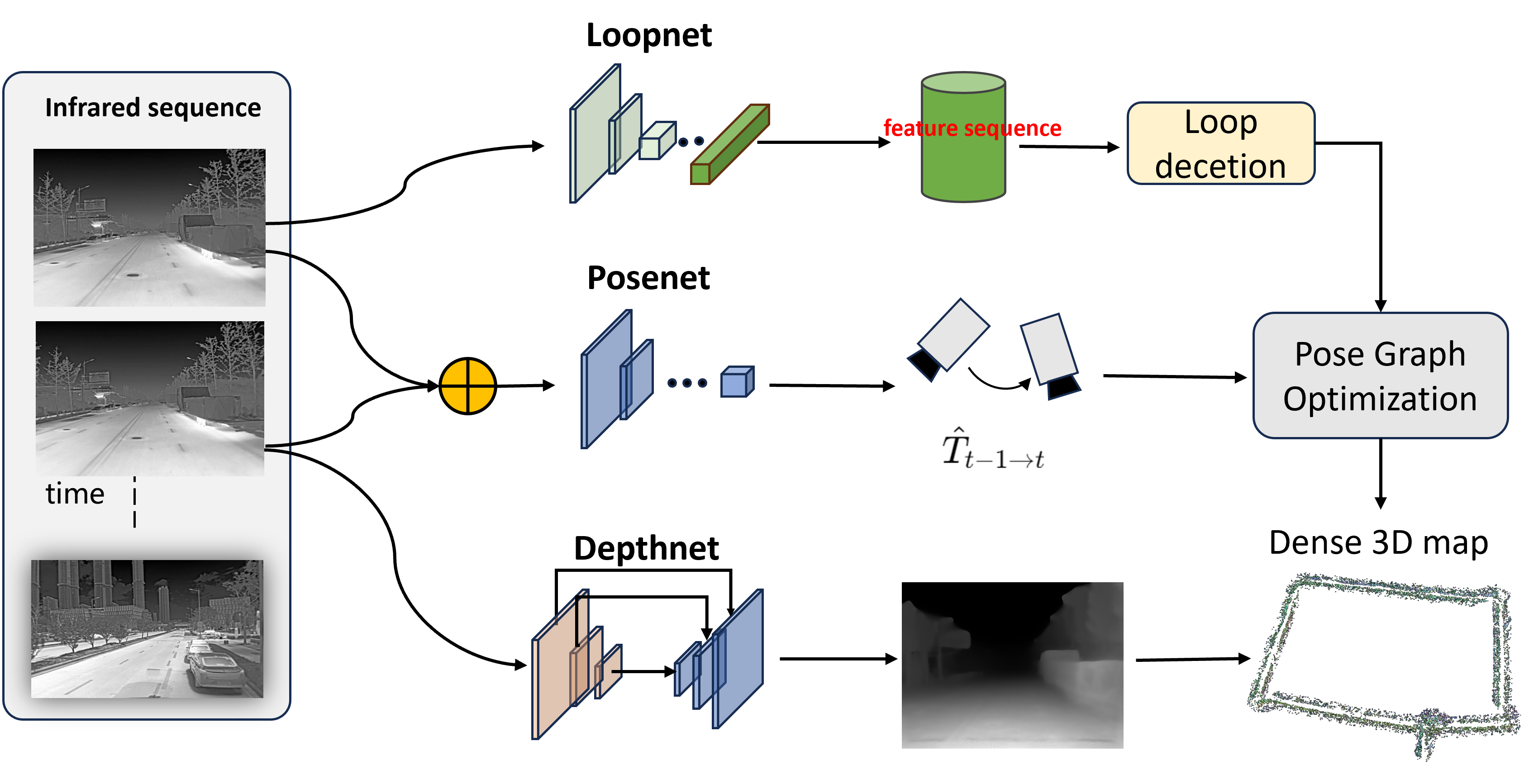}
    \caption{PoseNet computes relative poses and builds a pose graph from the image stream, while LoopNet extracts keyframe features for loop closure detection. Localization and mapping are achieved by combining the optimized pose graph and depth maps from DepthNet.}
\label{fig:deeptio_arch_test1}
\end{figure}

In recent years, significant research efforts have focused on integrating deep learning into visual SLAM. In the context of RGB images, novel view synthesis combined with self-supervised learning has been widely applied to achieve visual odometry and depth estimation, showing remarkable effectiveness. Works such as Zhou  \cite{zhou2017unsupervised}, Godard \cite{9009796}, and Zhang  \cite{zhang2023lite} have demonstrated the utility of self-supervised learning in RGB-based visual SLAM. By reducing reliance on manual annotations and utilizing geometric consistency constraints to learn features directly from unlabelled data, self-supervised learning not only enhances model adaptability but also improves the accuracy of visual odometry and depth estimation.

However, thermal images present unique challenges in visual SLAM due to difficulties in feature extraction, high noise levels, and loss of detail, making depth estimation and loop closure detection problematic. Deep learning, with its superior feature extraction and generalization capabilities, offers a potential solution to these challenges. Although monocular thermal visual odometry has seen some progress, such as in the work of  Shin et al. \cite{shin2021self,shin2022maximizing}, the approach primarily relies on local histogram processing of initial thermal images followed by self-supervised learning to estimate thermal visual odometry. While this method validates the feasibility of self-supervised learning in thermal imagery, its applicability is limited to indoor or small-scale scenes. The complete monocular thermal SLAM problem in outdoor, long-range thermal scenarios remains unresolved.

To address the challenges of large-scale, low-light environments, we propose DarkSLAM, a novel learning based monocular thermal SLAM framework. As shown in Fig. \ref{fig:deeptio_arch_test1}, our system first preprocesses raw thermal images using linear transformation and low-pass filtering. Self-supervised learning  \cite{9009796,8100183} is then employed to generate a heat consistency loss, reducing reliance on ground truth pose data and labeled datasets. In the learning model, the introduced Efficient Channel Attention (ECA) \cite{9156697} mechanism dynamically adjusts channel weights for better pose feature extraction, while Selective Kernel Attention (SKA) \cite{bao2022using}  enhances multi-scale depth feature fusion. We also integrate the Dino \cite{9709990} model into DepthNet to improve training convergence and depth estimation. Extensive experiments in large-scale outdoor thermal environments show that DarkSLAM siginificantly outperforms SC-SfM-Learner \cite{zhou2017unsupervised}, Shin et al. \cite{shin2022maximizing}, and Vödisch \cite{10.1007/978-3-031-25555-7_3} in both pose prediction and depth estimation.
Notably, incorporating the ECA mechanism reduced PoseNet’s Absolute Trajectory Error (ATE) by 38.5\% compared to SC-SfM-Learner. Meanwhile, the SKA mechanism not only enhances depth prediction accuracy but also effectively mitigates common issues related to depth prediction degradation.The Siamese network-based loop detection framework further improved loop closure success, optimizing the pose graph and delivering a robust thermal SLAM solution for precise localization and 3D mapping in challenging thermal conditions.
 
In summary, our key contributions are as follows:

\begin{itemize}
    \item We propose DarkSLAM, a novel self-supervised SLAM framework capable of learning poses, depths, and loop closures directly from monocular thermal images, that achieves large-scale thermal localization and mapping in complex lighting conditions.
    \item 
    By integrating Efficient Channel Attention and Selective Kernel Attention mechanisms into pose and depth estimation, we enhance the capability of pose and depth estimation while addressing common visual degradation issues in depth prediction.
    \item We introduce a contrastive learning-based loop closure detection module, enabling efficient loop closure and pose graph optimization in the thermal SLAM backend.
    \item We developed a thermal data collection platform ourselves and conducted extensive real-world experiments, demonstrating that DarkSLAM significantly outperforms existing solutions.
\end{itemize}

\section{RELATED WORK}
\subsection{Thermal SLAM}
Traditional thermal SLAM methods, which often extend classical visual SLAM algorithms, face significant limitations due to the unique properties of thermal images, such as low feature extraction precision and error accumulation that can lead to system failures. To address these challenges, researchers have proposed three main categories of solutions.
The first category involves integrating inertial sensors or radar with thermal SLAM  \cite{CHOI2023222}. While this approach improves localization, it requires precise sensor calibration and synchronization, and errors can still accumulate.
The second category aims to improve thermal SLAM by enhancing image processing before feature extraction\cite{9804793}. 
For instance, Wang et al\cite{10048516}. proposed novel thermal image processing and feature extraction strategies tailored to the unique thermal characteristics, achieving effective localization in dynamic environments with visual degradation.
The third category combines visual and thermal sensors for SLAM  \cite{10111061}, leveraging the strengths of both sensor types under varying lighting conditions. However, this approach introduces complexities such as image registration and data fusion, especially in environments with drastic lighting changes.
\begin{figure}
    \centering
    \includegraphics[width=0.95\columnwidth]{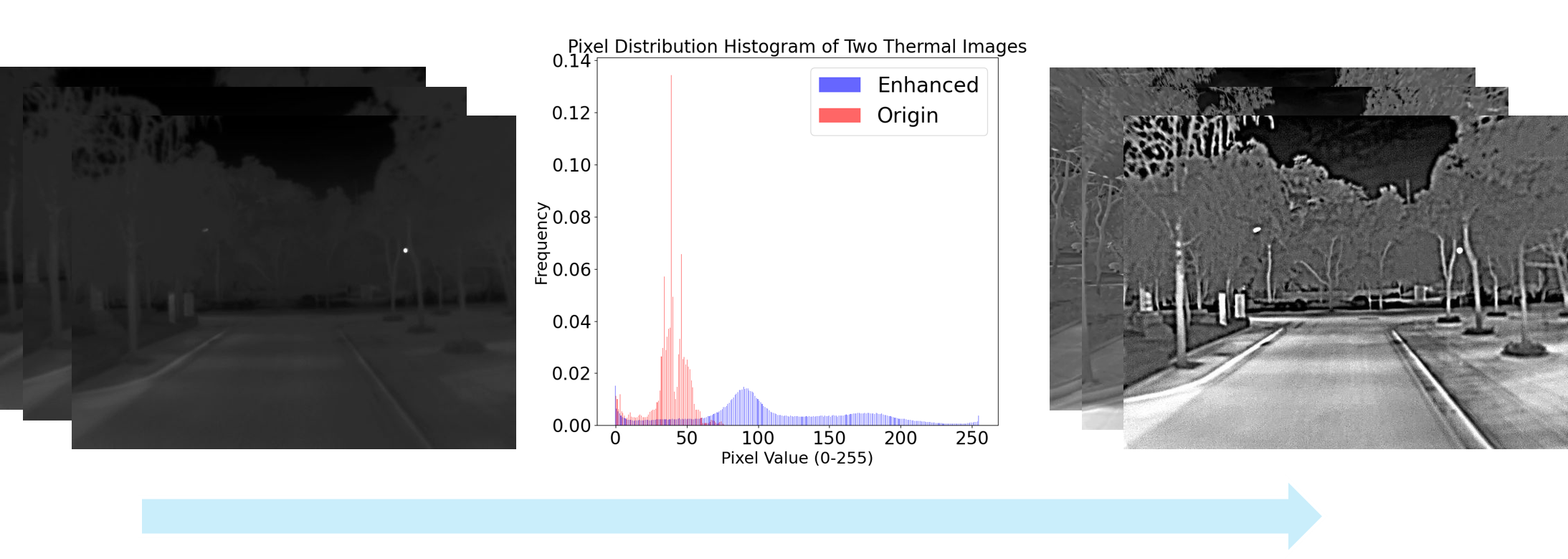}
    \caption{Comparison of the original thermal image (left) and the transformed image (right). Adjusting brightness and contrast enhances image features, clarifying dark details and highlighting grayscale differences. Filtering the detail layer reduces noise, resulting in a cleaner image.}
\label{fig:deeptio_arch_test2}
\end{figure}
\begin{figure*}
\centering
\includegraphics[width=2.0\columnwidth]{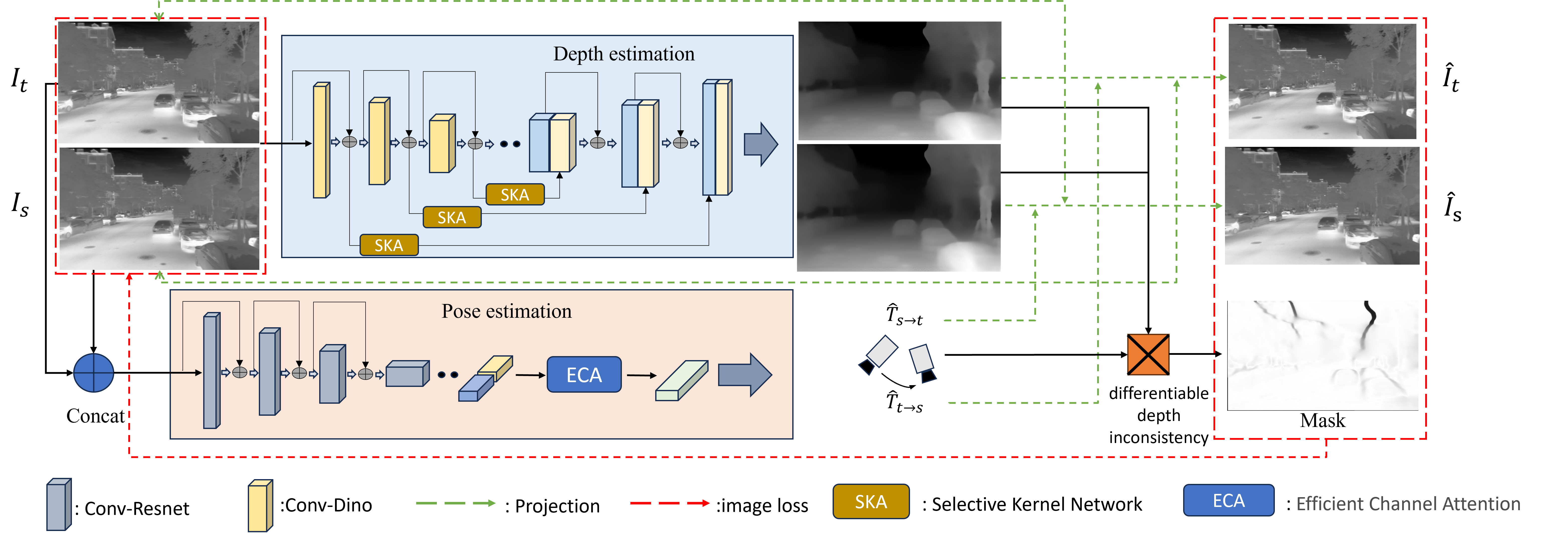}
\caption{In our proposed DarkSLAM framework, the pose and depth estimation modules adopt a self-supervised learning architecture. The predicted pose
and depth are used to warp the source image to generate new neighbor images, construct a mask to compute the image loss. }
\label{fig:deeptio_arch_train3}
\end{figure*}

\subsection{Deep Learning based Visual SLAM}
Self-supervised learning in RGB camera-based SLAM has gained popularity for depth and motion estimation from unlabeled image sequences, eliminating the need for costly ground-truth labels. A notable example is SC-Sfm-Learner \cite{zhou2017unsupervised}, which estimates depth and pose through temporal image reconstruction by warping one image to match another. Kim et al. \cite{kim2018multispectral} proposed using spatial RGB stereo reconstruction to train single-view depth estimation by aligning thermal with RGB images. Lu et al. \cite{lu2021alternative} introduced a cross-spectral reconstruction loss for single-view depth networks. Li et al. \cite{9047170} extended this framework by integrating LoopNet and pose graphs, significantly improving robustness and accuracy in RGB-based SLAM through temporal consistency and loop closure. However, adapting these methods to thermal images is challenging due to high noise, low feature precision, and spectral differences, which hinder long-range pose prediction and depth estimation in large-scale thermal environments. Thus, while RGB-based SLAM methods show promise, applying them to thermal imaging remains a complex task.

\section{Deep Learning Based Thermal SLAM}
In this section, we propose DarkSLAM, a novel deep learning-based thermal SLAM framework for large-scale localization and mapping using only monocular thermal images. The inherent challenges of thermal imaging, including feature scarcity, low contrast, and high noise, make it difficult for deep neural networks to directly extract meaningful features during training. To overcome these challenges, we first apply a transformation to raw thermal images to enhance contrast and detail while preserving temperature consistency. Next, we introduce a self-supervised learning framework to learn poses and depths from thermal images. We incorporate an Efficient Channel Attention (ECA) module  \cite{9156697} into PoseNet, which improves its sensitivity to differences between input image pairs and enhances pose feature extraction. For DepthNet, we use Dino-ResNet50  \cite{9709990} as the backbone encoder and integrate a Selective Kernel Network (SKA) module 
 \cite{bao2022using} in the depth decoder, mitigating depth feature degradation. Additionally, we propose a loop closure detection network based on a Siamese architecture, enabling reliable loop closure detection and pose optimization for thermal SLAM in large-scale environments.

\label{Thermal Images Consistent Enhancement  Framework  Overview}
\subsection{Thermal Image Enhancement}
Thermal imaging uses color coding to represent temperature distributions, but often suffers from low resolution, poor contrast, and lack of detail. While these characteristics can highlight temperature anomalies, they obscure scene details, making feature extraction difficult. To enhance feature extraction, raw thermal data must be processed by adjusting brightness and contrast and applying linear stretching techniques. First, the brightness range is stretched to improve texture contrast across temperature regions. A linear transformation then limits brightness within specific bounds. To further enhance image details, a Gaussian filter decomposes the image into background and detail layers. The background is smoothed to reduce noise, while the detail layer is enhanced to highlight key features. This processing improves contrast and provides more effective supervision signals for self-supervised learning, as shown in Fig. \ref{fig:deeptio_arch_test2}. 

\subsection{Self-Supervised Pose and Depth Learning from Thermal Images}
\label{Self-Supervised Framework  Overview}

Our self-supervised framework, as shown in Fig. \ref{fig:deeptio_arch_train3},DarKSLAM, enables depth and pose estimation from thermal images using novel view synthesis as the primary supervision signal, eliminating the need for ground-truth pose data. To tackle the challenges of thermal imaging, we integrate Efficient Channel Attention (ECA) into PoseNet for improved pose feature extraction and use Dino as DepthNet's backbone for faster convergence. Additionally, we incorporate a Selective Kernel Network (SKA) module to reduce depth degradation during joint training with PoseNet.

\subsubsection{Self-supervised Learning via Thermal Images Synthesis}

We select a target thermal image $I_{t}$ and an adjacent source view $I_{s}$ from a nearby time frame (e.g.,$t-1$ or $t+1$). DepthNet generates a per-pixel depth map ${D}_{t}$ from $I_{t}$, while PoseNet estimates the relative camera pose ${T}_{t\to s}$ using both $I_{t}$ and $I_{s}$. These outputs are used to inverse warp $I_{s}$ to align with $I_{t}$. A core component of our framework is a differentiable depth-based image renderer, which reconstructs the target view by sampling pixels from the source view $I_{s}$. This relies on ${D}_{t}$ and ${T}_{t\to s}$. The pixel transformation from target to source view is given by:
\begin{equation}
    p_s\sim K{T}_{t\to s}{D}_t(p_t)K^{-1}p_t.
\end{equation}
To reconstruct $\hat{I}_s(p_t)$, we use differentiable bilinear interpolation. This interpolates the values from the four neighboring pixels surrounding the projected pixel $p_{s}$. The formula is
\begin{equation}
    \hat{I}_{s}(p_{t})=I_{s}(p_{s})=\sum_{i\in\{t,b\},j\in\{l,r\}}w^{ij}I_{s}(p_{s}^{ij})
\end{equation}
where $w_{ij}$ are the interpolation weights. This interpolation is differentiable, ensuring smooth gradient propagation. This method allows for precise and stable image warping and supports end-to-end optimization, relying on projective geometry instead of directly learning the warping.

\subsubsection{Training Loss and Mask Generation}
The total loss function in the DarkSLAM framework is defined as:
\begin{equation}
L_{\mathrm{total}}=L_{\mathrm{rec}}+\lambda_{\mathrm{gc}}L_{\mathrm{gc}}+\lambda_{\mathrm{sm}}L_{\mathrm{sm}}
\end{equation}
This combines photometric consistency loss ($L_{rec}$)\cite{1284395}, geometric consistency loss ($L_{gc}$)\cite{bian2019unsupervised}, and edge-aware smoothness loss ($L_{sm}$)\cite{9009796} to train depth and pose estimation networks for self-supervised learning. The photometric consistency loss optimizes pixel-wise differences between synthesized and real images, computed using a weighting factor $\lambda$, and is expressed as:

\begin{multline}
    L_{\mathrm{rec}} = \frac{1}{|V|} \sum_{p \in V} \left( \lambda \| I_t(p) - \hat{I}_s(p) \|_1 \right. \\
    \left. + \frac{1 - \lambda}{2}\left( 1 - \mathrm{SSIM}(I_t(p), \hat{I}_t(p)) \right) \right)
\end{multline}
The geometric consistency loss minimizes depth inconsistencies between views to maintain scale consistency across the sequence. It is computed by aligning the synthesized depth $D_s^t$ for $I_{s}$, generated by $D_t$ and ${T}_{t\to s}$ with the rigid transformation, and comparing it with the interpolated depth $D_s^\prime$.
\begin{equation}
    L_{\mathrm{gc}}=\frac{1}{|V|}\sum_{p\in V}D_{\mathrm{diff}}(p),D_{\mathrm{diff}}(p)=\frac{|D_s^t(p)-D_s^\prime(p)|}{D_s^t(p)+D_s^\prime(p)}
\end{equation}
The edge-aware smoothness loss regularizes the depth map by incorporating image gradients, ensuring smoothness while preserving edge details. It is formulated as:
\begin{equation}
    L_{\mathrm{sm}}=\sum_p(e^{-\nabla I_t(p)}\cdot\nabla D_t(p))^2
\end{equation}
Dynamic masking is introduced to handle invalid pixels due to dynamic objects or occlusion. The self-discovered dynamic mask ($M_{sd}$) is based on depth inconsistency, defined as $M_{sd}=1-D_{\mathrm{diff}}$, where $\mathrm{diff}$ measures depth inconsistency between views. Additionally, an automatic mask ($M_{a}$) removes invalid pixels caused by static camera motion by comparing photometric losses between synthesized and reference images:
\begin{equation}
    M_a(p)=
\begin{cases}
1 & \mathrm{if}\|I_t(p)-\hat{I_s}(p)\|_1<\|I_t(p)-I_s(p)\|_1, \\
0 & \text{otherwise.} 
\end{cases}
\end{equation}

\subsubsection{ECA Guided Pose Network}
\label{sec:Network Details }
In our DarkSLAM, the PoseNet employs a ResNet-based \cite{he2016deep} encoder to extract image features from pairs of input images. These features are then fed into the PoseNet decoder to estimate the 6-DoF (Degrees of Freedom) pose. To better aggregate features from the two images, we incorporate the ECA (Efficient Channel Attention) module into the PoseNet decoder. Unlike traditional attention mechanisms, the ECA module does not require dimensionality reduction and upscaling operations, preserving the integrity of the original channel features. This allows the model to better exploit channel dependencies and enhance feature representation.

\subsubsection{Dino and SKA Enhanced Depth Network}
The DepthNet of our DarkSLAM is built on a ResUnet \cite{diakogiannis2020resunet} architecture. However, due to the limited availability of thermal datasets and the low contrast and high noise in thermal images, training DepthNet can be challenging, often leading to suboptimal performance. To address this issue, we incorporate the Dino (self-distillation with no labels) model \cite{9709990} as the encoder for DepthNet. Dino, trained on a large volume of images, has demonstrated exceptional performance across various vision tasks, including depth estimation. When used as the encoder in DepthNet, Dino effectively extracts features from thermal images, which are then utilized by the decoder for accurate depth prediction.

Additionally, due to the lack of texture information and low contrast in thermal images, DepthNet is prone to severe overfitting during self-supervised training. This overfitting often leads to the abandonment of depth predictions in low-texture regions, resulting in large black spots in the predicted depth maps. To mitigate this visual degradation, we introduce an SKA (Selective Kernel Attention) module \cite{bao2022using} in the upsampling feature fusion stage of DepthNet. The SKA module enables the network to better capture multi-scale features in complex thermal image spaces, preserving the network's ability to predict depth in weakly textured regions. Additionally, it enhances the network's perception and depth prediction capabilities for distant objects in thermal images.

\begin{figure}
    \centering
    \includegraphics[width=0.95\columnwidth]{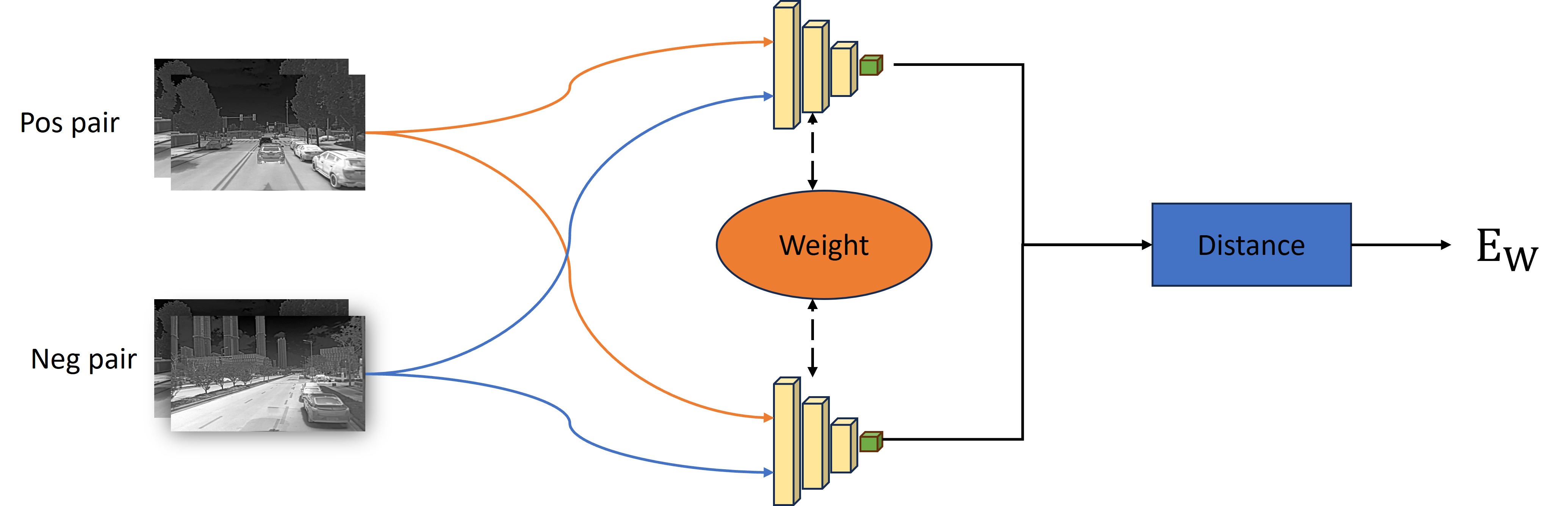}
    \caption{The figure illustrates the process of training LoopNet using a Siamese network. By minimizing the feature distance of positive samples and maximizing the feature distance of negative samples, we effectively enhance the loop closure detection performance. }
\label{fig:deeptio_validation}
\end{figure}

\subsection{Thermal Loop Closure Detection}
In thermal imaging, scene boundaries and features often appear blurred compared to RGB images, making traditional loop closure detection methods less effective. Hand-crafted visual features or texture-based methods struggle with the high similarity and low contrast inherent in thermal images. To overcome these challenges, we propose a Siamese network-based framework for loop closure detection using thermal images.

\subsubsection{LoopNet Architecture}
Our core network is based on ResNet18, trained using a contrastive learning approach that maximizes similarity between images of the same scene and minimizes it between different scenes. This enhances the network's ability to distinguish between similar and dissimilar scenes, even under the low contrast and noise typical of thermal images.The training framework is shown in Fig. \ref{fig:deeptio_validation}. 

We integrate this with a global loop closure detection and pose graph optimization system , combining PoseNet with our thermal loop closure detection network. PoseNet provides the relative pose information by calculating the transformation between adjacent images. During deployment, we construct a pose graph , incorporating both local and global connections, visual odometry (VO) estimation, and loop closure detection. Our loop closure module, a twin-network thermal detection system, converts each input frame into a feature vector, which is then compared to previous vectors using cosine similarity:

\begin{align}
\label{eq:reg_loss2}
\sin_{\cos}=\cos(\mathbf{f}_{\mathrm{current}},\mathbf{f}_{\mathrm{previous}})
\end{align}

When the cosine similarity exceeds a predefined threshold, PoseNet estimates the transformation between the corresponding images, updating the pose graph to ensure global trajectory accuracy and consistency.

\subsubsection{Training Details}
\label{sec:learn_pose}

In thermal loop detection, environmental temperature fluctuations and dynamic objects can cause significant variations in image brightness and contrast. To improve the model's accuracy and robustness, we applied image enhancement techniques during training. Specifically, we adjusted the brightness and contrast of positive sample pairs and randomly embedded patches from negative samples into the positive pairs, simulating real-world environmental changes. This strategy enhances the model's adaptability to varying imaging conditions. We also apply data augmentation (Fig. \ref{fig:deeptio_validation}) to improve model robustness. Given input samples $X_1$ and $X_2$, and model $G_W$, the contrastive loss is computed as:

\begin{align}
\label{eq:reg_loss}
     E_{W}=||G_W(X_1)-G_W(X_2)||
\end{align}

The loss function is defined as:
\begin{equation}
    \begin{split}
        L(W, (Y, X_1, X_2)^i) &= (1 - Y)L_G\left(E_W(X_1, X_2)^i\right) \\
        &\quad + YL_I\left(E_W(X_1, X_2)^i\right)
    \end{split}
\end{equation}

Here, $Y$ indicates whether $X_1$ and $X_2$ are from the same category ($Y = 0$) or different categories ($Y = 1$). $P$ is the total number of input data points, and $i$ is the current data point index. $L_G$ and $L_I$ are the loss functions for matching and non-matching pairs, respectively.

\subsection{Pose Graph Optimization}
In our DarkSLAM system, the backend optimization process begins with the initialization of a pose graph \cite{5979949}, where nodes represent thermal images and their corresponding global poses, and edges represent the relative poses between nodes. When a new thermal image frame arrives, PoseNet predicts the relative pose between the current and previous frames. Using the previous frame's global pose and the predicted relative pose, the global pose of the current frame is calculated. This new global pose is then added as a node in the pose graph, with the corresponding relative pose added as an edge connecting it to the previous node. When LoopNet detects a loop closure, PoseNet computes the relative poses between the current image and the images at the loop closure node. These relative poses are added as loop closure edges in the pose graph. The pose graph is then optimized through iterative refinement using the nonlinear optimization Ceres Solver, improving the accuracy of pose estimation and ensuring global trajectory consistency in complex thermal environments.
\begin{figure}[!ht]
    \centering
    \includegraphics[width=0.8\columnwidth]{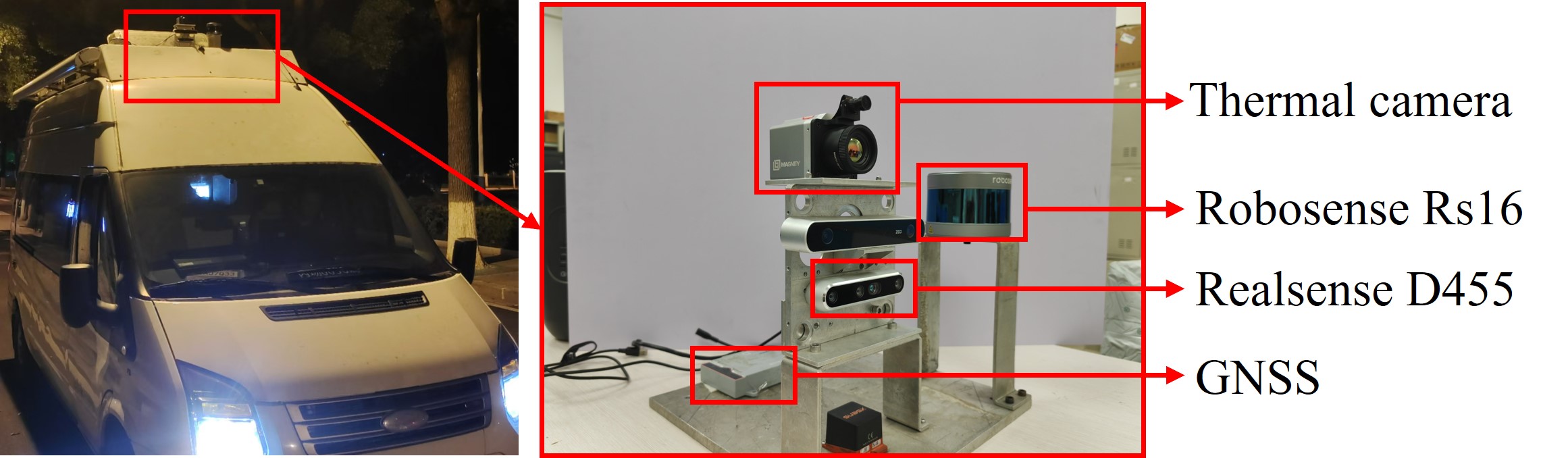}
    \caption{Experimental platform.}
\label{fig:rpe_validate_hallu}
\end{figure}

\section{EXPERIMENT}
\subsection{Experimental Setup}
\begin{figure*}[!ht]
    \centering
    \includegraphics[width=2\columnwidth]{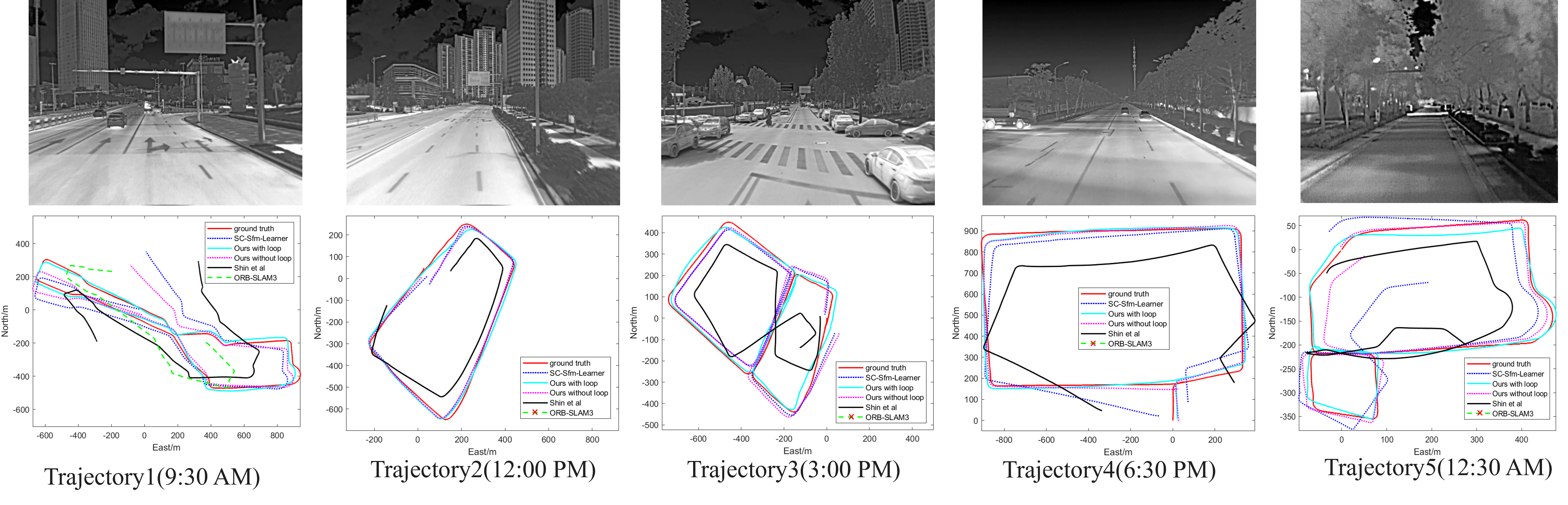}
    \caption{
    Comparison of trajectories from Shin et al., SC-Sfm-Learner,ORB-SLAM3 and our DarkSLAM (with and without loop detection) on trajectorys 1, 2, 3, 4, and 5, compared to ground truth trajectories.}
\label{fig:features_visualization4}
\end{figure*}

\subsubsection{Implementation Details}
In our training process, the self-supervised framework, consisting of DepthNet and PoseNet, and the thermal loop closure detection framework are trained separately. The encoder of DepthNet is initialized with a pre-trained DINO backbone, and its parameters are not frozen, allowing fine-tuning alongside PoseNet during training. 
We use the Adam optimizer \cite{Kingma2014AdamAM} with hyperparameters set to $\theta_1 = 0.9$ and $\theta2 = 0.999$, an initial learning rate of 0.0002, and a minibatch size of 4. Each epoch consists of 1,000 iterations, with a total of 100 epochs. All experiments are conducted on image sequences captured by a monocular thermal camera, maintaining the original resolution of 640x480 pixels during training. For the Siamese network in the thermal loop closure detection framework, we select positive and negative samples augmented with data for training. The Adam optimizer is used with a learning rate of 0.0001, a batch size of 32, and each epoch consists of 1,000 iterations, over a total of 100 epochs. 

\subsubsection{Data Collection}
Thermal data was captured using a MAGNITY thermal camera at 10 Hz and a resolution of 640x480 pixels, while GNSS data from the SinoGNSS system served as ground truth for trajectory evaluation. The data was collected with a vehicle-mounted platform across 11 urban road sections at various times of day, resulting in 85,091 thermal images for training. An additional 20,495 images from 5 road sections were used for testing, forming the Largescale Thermal Dataset. RGB images were synchronized using a Realsense D455 camera, and depth data was captured with a RoboSense RS16 LiDAR.


\begin{figure}
    \centering
    \includegraphics[width=0.8\columnwidth]{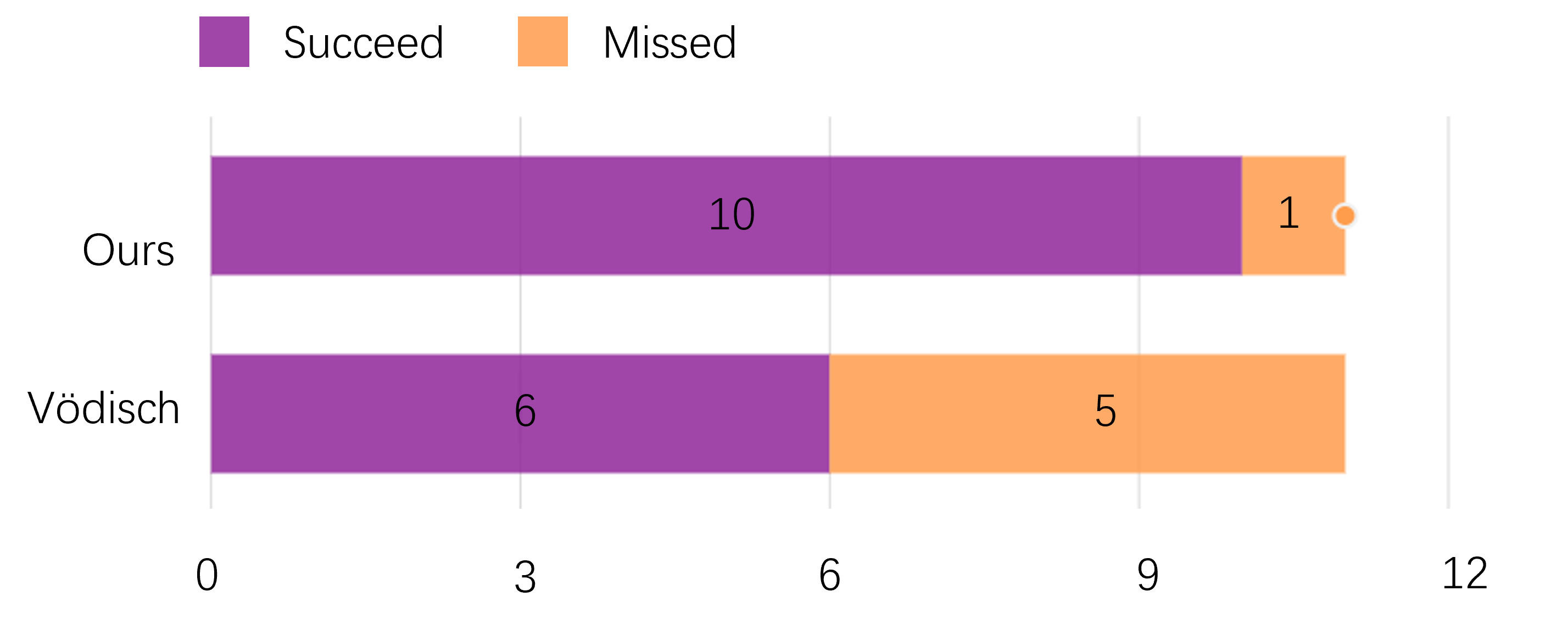}
    \caption{
    Comparison of our method and Vödisch  \cite{10.1007/978-3-031-25555-7_3}'s LoopNet performance at 11 loop closures.}
    \label{fig:deeptio_validation1}
\end{figure}

\begin{table*}[]
\caption{
Comparison of PoseNet ATE and RPE values for SC-Sfm-Learner \cite{zhou2017unsupervised}, Shin et al. \cite{shin2022maximizing}, ORB-SLAM3\cite{9440682}, and our DarkSLAM (with and without Efficient Channel Attention) across five test trajectories.}
\label{tab:table1}
\centering
\resizebox{\linewidth}{!}{%
\begin{tabular}{l|cc|cc|cc|cc|cc} 
\toprule
Method               & \multicolumn{2}{c|}{SC-Sfm-Learner} & \multicolumn{2}{c|}{Shin et al.} & \multicolumn{2}{c|}{ORB-SLAM3} & \multicolumn{2}{c|}{Ours without ECA} & \multicolumn{2}{c}{Ours} \\ 
\midrule
Error                & ATE (m) & RPE (deg) & ATE (m) & RPE (deg) & ATE (m) & RPE (deg) & ATE (m) & RPE (deg) & ATE (m) & RPE (deg) \\ 
\midrule
Trajectory1          & 111.014 & 0.280     & 271.550 & 0.309     & 308.773 & 0.599     & 281.936 & 0.267     & \textbf{74.744}  & \textbf{0.252} \\
Trajectory2          & 48.782  & 0.303     & 118.640 & 0.324     & -       & -         & 51.345  & 0.319     & \textbf{47.103}  & 0.305 \\
Trajectory3          & 48.773  & 0.306     & 157.609 & 0.294     & -       & -         & 44.683  & 0.306     & \textbf{40.953}  & \textbf{0.290} \\
Trajectory4          & 89.265  & 0.253     & 221.680 & 0.265     & -       & -         & 82.554  & 0.256     & \textbf{27.750}  & \textbf{0.244} \\
Trajectory5          & 67.573  & 0.338     & 140.085 & 0.381     & -       & -         & 81.813  & 0.360     & \textbf{34.100}  & 0.356 \\ 
\midrule
Average              & 73.081  & 0.296     & 181.913 & 0.315     & 308.773 & 0.599     & 108.466 & 0.302     & \textbf{44.930}  & \textbf{0.289} \\ 
\bottomrule
\end{tabular}
}
\end{table*}

\subsubsection{Efficiency Analysis}
During the prediction phase, our algorithm encompasses several steps, including histogram processing, pose estimation, depth estimation, loop closure detection, pose graph construction and optimization, and map building.
For input images of size $H\times W$, PoseNet exhibits a computational complexity of $O(C^2\times K^2\times H\times W\times L)$, where $C$ is the number of convolutional channels, $K$ is the kernel size, and $L$ is the network depth. DispResNet and LoopNet in comparison, have a complexity of $O(C_{\max}^2\times H\times W)$, where $C_{\max}$ is the maximum number of channels in the network.
In practical validation, our DarkSLAM system, running on a 24GB NVIDIA RTX 3090 GPU with an input image resolution of 640×480 pixels, achieved a pose estimation speed of approximately 12 frames per second during prediction, including loop closure detection, with a memory usage of around 3,162 MB. When loop closure detection was disabled, the pose estimation speed increased to 17 frames per second, while memory usage decreased to approximately 3,061 MB. Depth estimation required about 6,614 MB of memory while maintaining a consistent prediction speed of 17 frames per second. These results demonstrate that DarkSLAM delivers real-time pose and depth processing, ensuring efficiency in practical applications.
 

\begin{figure}[!ht]
    \centering
    \includegraphics[width=1.0\columnwidth]{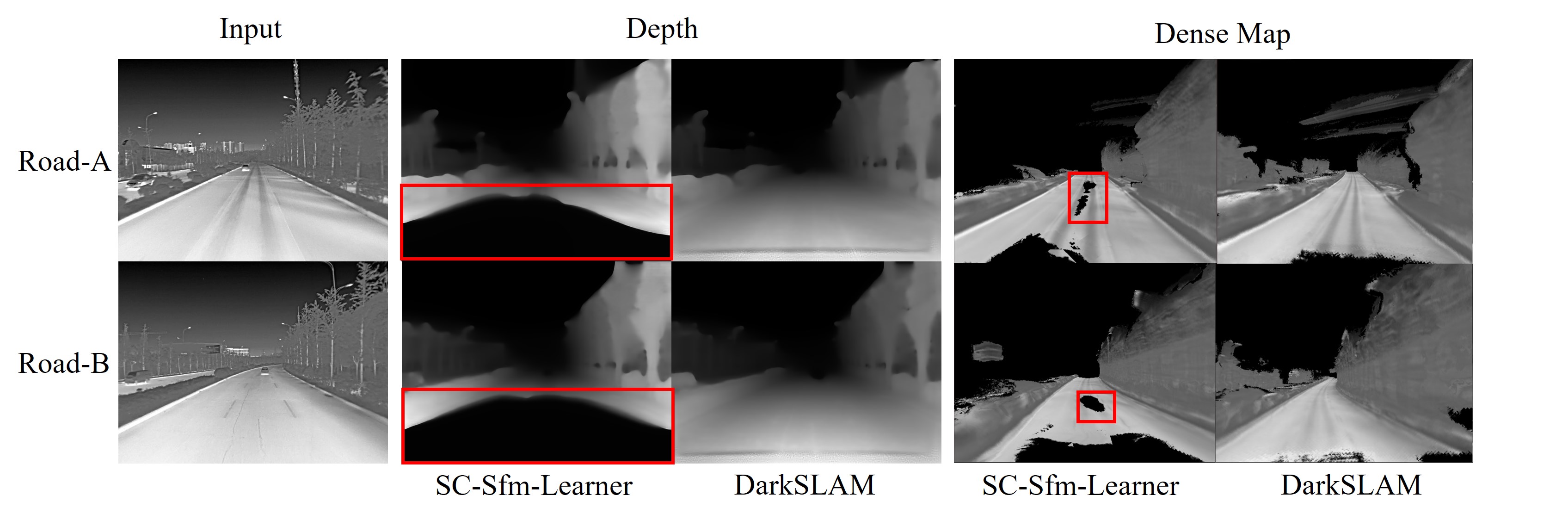}
    \caption{Comparison of DepthNet's depth maps with 3D reconstructions from SC-Sfm-Learner(left) and DarkSLAM(right). Red boxes indicate areas of visual degradation affecting 3D quality.}
\label{fig:features_visualization2}
\end{figure}

\subsection{Large-scale Thermal Localization and Mapping}
\subsubsection{Pose Estimation Evaluation}
We evaluated pose prediction on five thermal urban road sections that were excluded from the training set: trajectory1 (3,491 images), trajectory2 (6,473 images), trajectory3 (2,460 images), trajectory4 (4,751 images), and trajectory5 (3,320 images). 
 We evaluated the trajectory performance of Shin et al. \cite{shin2022maximizing}, SC-Sfm-Learner \cite{zhou2017unsupervised}, ORB-SLAM3\cite{9440682} and our DarkSLAM framework, both with and without loop detection and pose optimization. Shin et al.’s PoseNet and DepthNet shared a ResNet50-based encoder, while SC-Sfm-Learner's PoseNet and DepthNet used the same encoders as DarkSLAM, ResNet18 and ResNet50, respectively.
To ensure a fair comparison, all methods were trained and tested on thermally enhanced images to address challenges like high noise, low contrast, and inconsistent brightness in the original thermal data.Additionally, ORB-SLAM3 was tested using RGB images collected from the RealSense D455 camera. The trajectory generation results are shown in Fig. \ref{fig:features_visualization4}.

\begin{figure}[!ht]
    \centering
    \includegraphics[width=0.8\columnwidth]{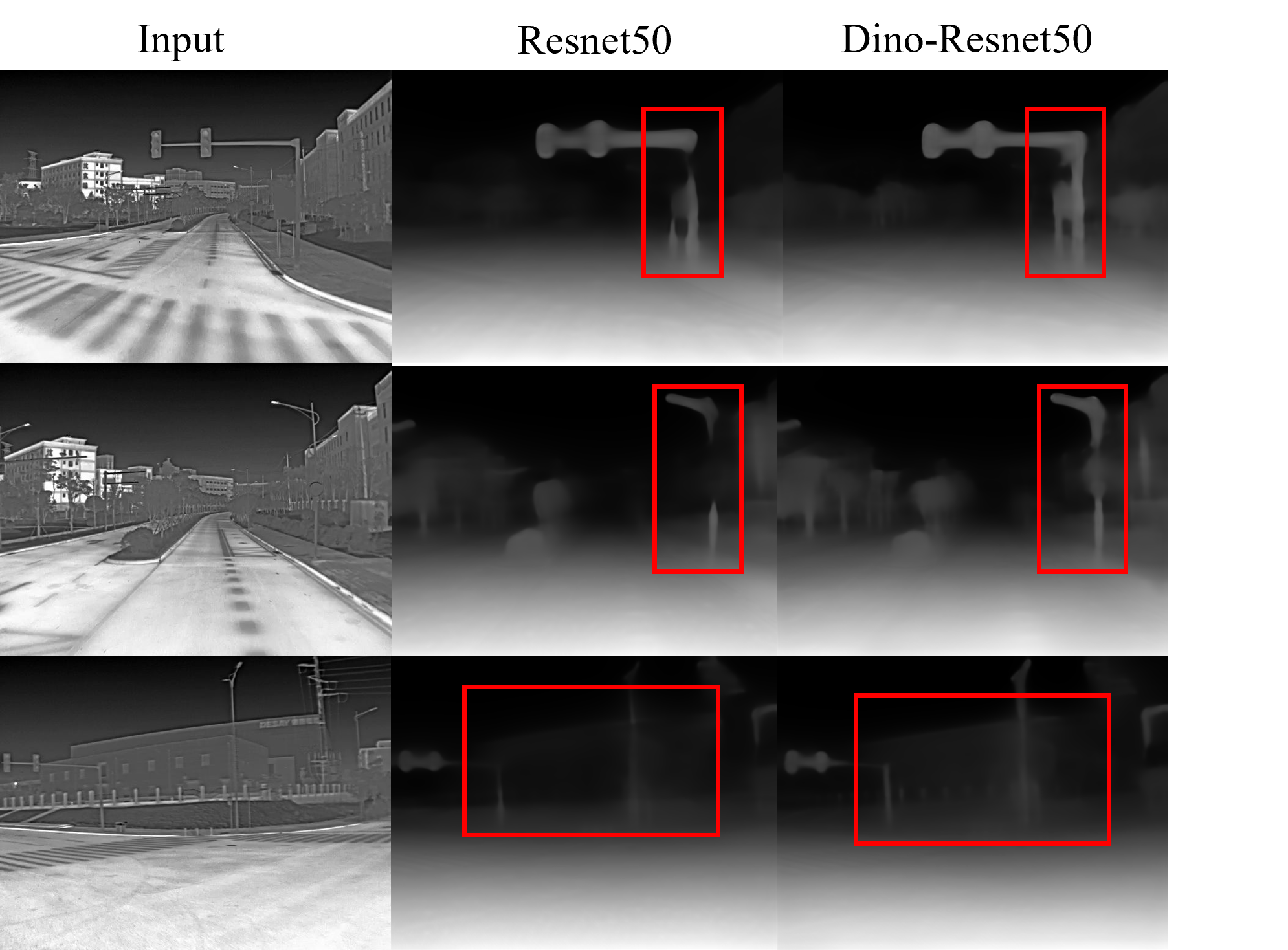}
    \caption{Comparison of DepthNet using ResNet50 and Dino-ResNet50 \cite{9709990} as encoders, comparing the depth estimation performance from left to right }
\label{fig:rpe_validate_hallu2}
\end{figure}


\subsubsection{Loop-closure Detection Performance}
To evaluate the performance of our thermal loop detection module, we compared it with the method proposed by Vödisch \cite{10.1007/978-3-031-25555-7_3}. Loop detection accuracy is highly sensitive to the feature similarity threshold. Raising the threshold typically increases both false positives and false negatives. Striking the right balance is critical, as a higher threshold may falsely identify loops, while a lower threshold may miss true loops. To address this, we implemented a dynamic threshold adjustment strategy, gradually tuning the threshold across different scenarios until false positives were minimized, thereby identifying the optimal threshold. Using this threshold, we compared the detection success rates and false negative rates of both methods across various routes. Key frames were extracted every 15 frames, and after detecting a loop, detection was suspended for the next 150 frames. The routes (trajectories 1, 2, 3, 4, and 5) contained 2, 1, 4, 1, and 3 loops, respectively. Our method successfully detected 2, 1, 3, 1, and 3 loops, while Vödisch's method detected 1, 1, 2, 0, and 2 loops. Overall, our loop detection success rate improved by 66.6\% compared to Vödisch’s method, as shown in Fig. \ref{fig:deeptio_validation1}. This highlights the effectiveness of our dynamic threshold adjustment and enhanced feature extraction, resulting in more reliable loop detection in challenging thermal environments.

\subsection{ \textit{Ablation Study}  }

\begin{table}[t]
    \caption{
    VARIOUS METRICS ARE USED TO EVALUATE THE OUTPUT OF
DEPTHNET,AND A1, A2, AND A3 REPRESENT THRESHOLD ACCURACY
(\textless1.25, \textless1.25², \textless1.25³)}
\label{tab:table2}
\centering
\resizebox{\linewidth}{!}{%
\begin{tabular}{l|l|llllll} 
\toprule
Method                            & Sequence             & Abs Rel$\downarrow$         & Sq Rel$\downarrow$           & RMSE$\downarrow$             & a1$\uparrow$              & a2$\uparrow$              & a3$\uparrow$               \\ 
\midrule
\multirow{4}{*}{SC-Sfm-Learner}   & path1                & 0.2930          & 10.3149          & 20.5012          & 0.6387          & 0.8443          & 0.9152           \\
                                  & path2                & 0.4323          & 14.5034          & 27.7229          & 0.3934          & 0.6955          & 0.8240           \\
                                  & path3                & 0.5201          & 23.3934          & 21.5819          & 0.2754          & 0.8296          & 0.9259           \\
                                  & average              & 0.4151          & 16.0706          & 23.2687          & 0.4358          & 0.8119          & 0.8884           \\ 
\midrule
\multirow{4}{*}{Ours without SKA} & path1                & 0.2813          & 7.4723           & 15.8498          & 0.6384          & 0.8343          & 0.9312           \\
                                  & path2                & 0.4451          & 20.2038          & 21.2963          & 0.4790          & 0.7138          & 0.8404           \\
                                  & path3                & 0.4829          & 16.1368          & 20.8724          & 0.2163          & 0.7886          & 0.9229           \\
                                  & average              & 0.4031          & 14.6043          & 19.3395          & 0.4446          & 0.7789          & 0.8981           \\ 
\midrule
\multirow{4}{*}{Ours}             & path1                & 0.2749          & 8.1188           & 17.4654          & 0.6558          & 0.8502          & 0.9241           \\
                                  & path2                & 0.2799          & 4.0647           & 11.6565          & 0.5998          & 0.7880          & 0.9045           \\
                                  & path3                & 0.5188          & 20.1463          & 20.4462          & 0.2683          & 0.8315          & 0.9180           \\
                                  & average              & \textbf{0.3579} & \textbf{10.7766} & \textbf{16.5227} & \textbf{0.5080} & \textbf{0.8232} & \textbf{0.9155}  \\ 
\bottomrule
\end{tabular}
}
\end{table}

\subsubsection{Thermal Scene Reconstruction Performance}
During depth estimation, DepthNet experienced significant feature degradation over time due to the high similarity of thermal image features. This degradation resulted in numerous "black spots" in the predicted depth maps (Fig. \ref{fig:features_visualization2}), leading to incomplete depth maps and blank areas in the final dense 3D reconstructions, ultimately affecting mapping accuracy. The introduction of the Selective Kernel Attention (SKA) mechanism \cite{10023305} allowed the network to capture multi-scale spatial features more effectively, greatly reducing the feature degradation issue. To validate this improvement, we selected two degraded thermal sections for depth prediction and dense mapping comparison experiments. We compared depth maps generated by SC-Sfm-Learner \cite{zhou2017unsupervised} with those produced by DarkSLAM, analyzing their impact on thermal 3D reconstruction. The estimated depth maps from DarkSLAM were input into the InfiniTAM-V3 \cite{InfiniTAM_ISMAR_2015} framework to generate dense thermal point clouds. The results, shown in Fig. \ref{fig:features_visualization2}, illustrate the differences in point clouds generated by various depth predictions and demonstrate the improvements in the final 3D reconstructions.

\subsubsection{Ablation Study on DepthNet} 
We enhance DepthNet by integrating Dino-ResNet50 for initialization and the Selective Kernel Attention (SKA) mechanism \cite{10023305} to improve depth prediction performance and prevent depth degradation. Pre-training with Dino-ResNet50 enabled our network to extract more robust features from noisy, low-contrast thermal images, thereby increasing depth prediction accuracy (see Fig.\ref{fig:rpe_validate_hallu2}). We evaluated three DepthNet variants: the original SC-SfMLearner DepthNet, DepthNet with Dino-ResNet50 (without SKA), and DepthNet with both Dino-ResNet50 and SKA. Using LiDAR data as depth ground-truth and evaluation metrics such as Absolute Relative Error, Squared Relative Error, RMSE, and depth accuracy thresholds, the results in Table \ref{tab:table2} demonstrate that our proposed depth model integrating Dino-ResNet50 and SKA achieved superior depth accuracy and robustness over other settings, especially in low-feature and distant thermal scenes. These findings confirm the effectiveness of the enhancements of Dino and SKA in challenging environments.


\subsubsection{Ablation Study into ECA Module} 
We specifically evaluated the impact of the Efficient Channel Attention (ECA) \cite{9156697} on pose prediction by comparing DarkSLAM’s PoseNet with and without ECA. For a broader comparison, we also included PoseNets from Shin et al. \cite{shin2022maximizing} and SC-Sfm-Learner \cite{zhou2017unsupervised}. This allowed us to assess the effectiveness of ECA in improving thermal pose prediction. While Shin et al. used ResNet50, SC-Sfm-Learner employed ResNet50 for DepthNet and ResNet18 for PoseNet, and DarkSLAM utilized Dino-ResNet50 \cite{9709990} for DepthNet and ResNet18 for PoseNet. We tested on trajectories 1–5 from the Largescale Thermal Dataset, comparing Absolute Trajectory Error (ATE) and Relative Pose Error (RPE). ECA significantly improved pose prediction by enhancing the model’s attention to relevant features in noisy thermal images. DarkSLAM with ECA reduced ATE by 38.5\% and RPE by 2\% which can show in Table \ref{tab:table1} , outperforming SC-Sfm-Learner in both positional and rotational accuracy.


\section{CONCLUSIONS}
In this work, we present DarkSLAM, a deep learning-base monocular thermal SLAM system designed for large-scale, complex thermal environments. Leveraging self-supervised learning, DarkSLAM provides accurate pose estimation and depth mapping without the need for ground truth data. We incorporate attention mechanisms and the Dino model to address visual degradation in thermal images, while a siamese network-based contrastive learning framework enhances loop detection, significantly boosting overall performance.Future work will focus on improving loop detection in varying thermal conditions (e.g., day-to-night transitions) and enhancing dynamic object detection for more precise pose prediction.
However, there are several limitations. The large number of parameters in PoseNet, DepthNet, and LoopNet makes real-time deployment on resource-constrained edge devices challenging. Additionally, our loop optimization currently only refines the pose graph, not the depth predictions, limiting the generation of high-precision maps for applications like autonomous driving. While effective in static environments, the system struggles with dynamic objects, leading to degraded map accuracy when object density is high.

\printbibliography  

\end{document}